\documentclass[twoside,11pt]{article}

\usepackage{jmlr2e}
\usepackage{listings}
\usepackage{url}
\usepackage{subfigure}
\usepackage{setspace}

\jmlrheading{0}{0000}{0-00}{0/00}{00/00}{}

\ShortHeadings{Multi-Label Classification toolbox}{Kimura, Sun and Kudo}
\firstpageno{1}

\begin{document}

\title{MLC Toolbox: A MATLAB/OCTAVE Library for Multi-Label Classification  }

\author{\name Keigo Kimura \email kkimura@main.ist.hokudai.ac.jp \\
       \addr Department of Computer Science\\
       Hokkaido University
       \AND
       \name Lu Sun  \email  sunlu@main.ist.hokudai.ac.jp\\
        \addr Department of Computer Science\\
        Hokkaido University
        \AND
        \name Mineichi Kudo  \email  mine@main.ist.hokudai.ac.jp\\
        \addr Department of Computer Science\\
        Hokkaido University}

\editor{John Doe}

\maketitle

\begin{abstract}
Multi-Label Classification toolbox is a MATLAB/OCTAVE library for Multi-Label Classification (MLC). There exists a few Java libraries for MLC, but no MATLAB/OCTAVE library that covers various methods. This toolbox offers an environment for evaluation, comparison and visualization of the MLC results. 
One attraction of this toolbox is that it enables us to try many combinations of feature space dimension reduction, sample clustering, label space dimension reduction and ensemble, etc. 

\end{abstract}


\section{Multi-Label Classification and Libraries}

Multi-Label Classification (MLC), a problem which allows an instance to have more than one label at the same time,  becomes popular since the problem fits real applications more than traditional single-label classification. Since combinations of labels must be considered to solve MLC, MLC gives many challenging problems and invokes many works to solve them. 

There are already a few  libraries for MLC available. Mulan is the most popular Java library for MLC developed by  \cite{tsoumakas2011mulan}. This library provides not only many MLC methods but also many MLC datasets.  MEKA is another popular Java library for MLC developed by \cite{read2016meka}. MEKA provides GUI and is specialized for a family of methods called Classifier Chain (CC) (\cite{read2011classifier}). Both  Mulan and MEKA are based on Weka, that is, a popular classification Java library developed by \cite{hall2009weka}. Thus, both of Mulan and MEKA inherit methods implemented on Weka such as decision trees or SMO algorithm for SVMs. Scikit-multilearn (\cite{2017arXiv170201460S}) is a recently developed library on python and its machine learning library, scikit-learn  \cite{scikit-learn}. This library can utilize some methods of scikit-learn and have an interface of MEKA, however, available methods are limited. On the other hand, recent many implementations of MLC methods are based on MATLAB/OCTAVE. For example, as summarized on \cite{de2009tutorial}, LAMDA group\footnote{\url{http://lamda.nju.edu.cn/Data.ashx}} or Prof. ML-Zhang.\footnote{ \url{http://cse.seu.edu.cn/people/zhangml/Resources.htm}} This is because MATLAB/OCTAVE is easier to implement and run codes. In addition, MATLAB/OCTAVE can operate matrix or vector functions faster than those in Java in general. This is also helped by MEX technique that incorporates with C++ codes. However, there is no libraries for MLC on MATLAB/OCTAVE. This is our motivation to have developed this library. 

\section{MLC toolbox}
\begin{figure}[t]
	\includegraphics[width=150mm]{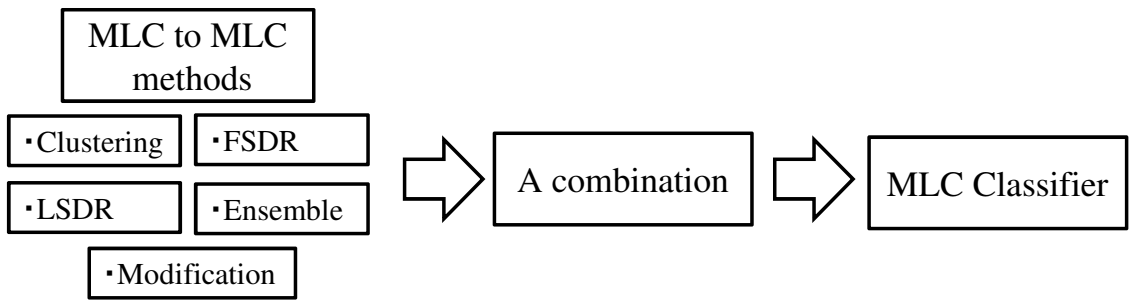}
	\caption{An illustration of MLC toolbox. MLC toolbox can combine methods convert a given MLC to more than one smaller or simpler MLCs (MLC to MLC methods) in any order. }
	\label{fig}
\end{figure}
This toolbox, a MATLAB/OCTAVE library for MLC, offers an easy implementation of various MLC methods and enables to compare them with each other easily.  There are many publicly available feature extraction methods and label space dimension reduction methods for MLC in MATLAB/OCTAVE codes. However, some clustering based methods or ensemble methods such as HOMER or RAkEL are not publicly available yet. This is probably because  
 MATLAB/OCTAVE is specialized for vector/matrix operations. This specialty might have produced two different groups: one is the group using Mulan or MEKA and the other using MATLAB/OCTAVE. This separation problem should be resolved to enable comparison over these two groups.  MLC toolbox provides one solution for this goal.  MLC toolbox provides some methods not familiar with MATLAB/OCTAVE environments. 
 
The most attractive point of MLC toolbox is that you can handle many combinations of MLC methods. As known well,  many MLC methods convert a given  MLC problem into more than one simpler or smaller MLC problems. For example, CBMLC  (\cite{nasierding2009clustering}) uses a clustering and produces smaller MLC problems. Feature extraction methods can also be considered as converting a problem into a smaller problem w.r.t. the size of features.  
These can be easily combined together in any order. For example, SLEEC (\cite{Kar2015sparse}) combines  clustering and  feature extraction. 
These combinations exist in myriad. 
MLC toolbox enables to try much more combinations of MLC methods in any orders more easily (See Fig. \ref{fig}).  

MLC toolbox is a library on MATLAB/OCTAVE and is flexible enough for allowing you to add your own method. Unfortunately, some dependencies exist. For example,it requires LIBLINEAR \cite{fan2008liblinear} and LIBSVM \cite{chang2011libsvm} for fast computation.  Nevertheless, this toolbox is OS-agnostic and thus offhand. 

MLC toolbox focuses only on supervised MLC. Thus, semi-supervised, incremental and missing labels are not supported. This is a drawback of this library against MEKA. 

Table \ref{tb:comparison} summarizes the comparison of MLC methods available.\footnote{LAMDA and ML-Zhang are collections of MLC methods.} The category of MLC methods are mostly based on \cite{de2009tutorial} but with some modifications: Ranking via Single-Label learning (RSL), Binary Relevance (BR), Label-Combination (LC), Pair-Wise comparison (PW), Dependencies, Neural-Network (NN), Deep Neural Network (DNN), Support Vector Machines (SVMs), Meta-Learners, Imbalance, Feature Space Dimension Reduction (FSDR), Label Space Dimension Reduction (LSDR), Clustering-based method (Clustering), Thresholding strategies and algorithms (Thresholding)

\begin{table}[t]
	\centering
	\caption{Comparison of Libraries} 
	\label{tb:comparison}
{\small	
	\begin{tabular}{ccccccc}
		\hline
		&Mulan &MEKA &LAMDA &ML-Zhang&scikit-multilearn& \textbf{MLC toolbox} \\ \hline
		Language & Java & Java & MATLAB &MATLAB&python&  MATLAB \\
		\cline{2-7} & \multicolumn{6} {c} {Available  Methods} \\ \cline{2-7}
		\textbf{Combination}&  &   & & & &\checkmark \\ 
		RSL          & \checkmark & \checkmark & & & & \checkmark\\
		BR          & \checkmark &\checkmark &  & & \checkmark &\checkmark \\
		LC            & \checkmark & \checkmark & & &\checkmark &\checkmark \\
		PW   & \checkmark &                   & & & &\checkmark \\
		Dependencies & \checkmark & \checkmark &\checkmark & \checkmark & \checkmark& \checkmark \\
		NN         & \checkmark &  & \checkmark&\checkmark& \checkmark&\checkmark		 \\
		Deep NN        & \checkmark & & & & & \\
		SVMs       &  & & &\checkmark& &\checkmark \\
		Meta-Learners & \checkmark & \checkmark &  & & \checkmark &\checkmark \\
		Imbalance  &   &   &  &\checkmark &  & \checkmark \\ 
		FSDR       &   &    & \checkmark & \checkmark &  & \checkmark\\
		LSDR    &  &   &    &  & & \checkmark \\ 
		Clustering &  & &  &  & \checkmark &\checkmark \\
		Thresholding & \checkmark & \checkmark &  & & \checkmark  & \checkmark \\
		\hline
	\end{tabular}
}	
\end{table}

\section{How to Use MLC toolbox}
\subsection{Set Up}
MLC toolbox needs MATLAB and the statistical and machine learning toolbox provided by Mathworks.
After downloading MLC toolbox, MLC toolbox requires only to compile some mex functions by running compileMEXfunctions.m on the main directory folder. Once you add the path to the all folders in MLC toolbox, you can run every program of MLC toolbox. 
\subsection{Running Sample Codes}
As an example, we describe how to run  Sample.m (a sample code to run MLC methods) briefly.
\\ \textbf{Dataset}: 
\begin{lstlisting}[basicstyle=\ttfamily\footnotesize]
datasetname='scene';  ## set the dataset name
numCV=5;  ## numCV=3 or numCV=10
\end{lstlisting}
\textbf{Method}:  
\begin{lstlisting}[basicstyle=\ttfamily\footnotesize]
method.name={'PCA','CBMLC','rCC'}; 
\end{lstlisting} 
As we stated above, MLC toolbox can handle combinations of MLC methods. MLC toolbox combines components in this order.
\\ \textbf{Parameters}: This library does not provide an efficient way to set parameters. This is because it is mostly impossible to set parameters on myriad combinations of methods. Thus, this library needs parameter files to set parameters for each MLC method. For sample codes, as an example, we give a parameters file (setMethodParameter.m) for each method. For example, to change parameter of PCA, change values of SetPCAParameter.m: 
\begin{lstlisting}[basicstyle=\ttfamily\footnotesize]
function[param]=SetPCAParameter(~)
param.dim=300;  or  param.dim='numF*0.5';  ## numF is the number of features
\end{lstlisting}  
The latter definition refers to the dataset information (reducing 50\% of feature dimensions). 
\\ \textbf{Base classifier}: 
Most of all methods except algorithm adopting methods calls traditional binary/multi-class classifiers to solve MLC. We call base classifiers by 
\begin{lstlisting}[basicstyle=\ttfamily\footnotesize]
method.base.name='linear_svm';  ## 'ridge','svm','knn' are available
method.base.param.svmparam='-S 2 -q'; ##parameters
\end{lstlisting}  
\textbf{Thresholding}: threshold method and parameters to obtain discrete results by
\begin{lstlisting}[basicstyle=\ttfamily\footnotesize]
method.th.type='Scut';  ## 'Rcut','Pcut' are also available
method.th.param=0.5; ##parameters
\end{lstlisting} 
See methods on \cite{tang2009large}.
\subsection{Implementing Your Codes in MLC toolbox}
Users do not have to understand the whole part of MLC toolbox to implement your own codes. For implementing a new method, say Newmethod, MLC toolbox requires to implement Newmethod\_train.m for training, Newmethod\_test.m for testing and SetNewmethodParameter.m for parameter setting (not necessary but recommended).  More detailed instruction information is given on a tutorial document. 
\section{Summary}
This MLC toolbox is the first toolbox that contains the largest set of methods and allows many combinations of them for MLC on MATLAB/OCTAVE. The MLC toolbox is available free for academic purposes. The software, documents, and tutorial slides can be available at \url{https://github.com/KKimura360/MLC_toolbox}.  We welcome your feedback and contribution to this toolbox.

\bibliography{MLC}

\end{document}